\documentclass[letterpaper, 10 pt, conference]{ieeeconf}  

\usepackage[utf8]{inputenc}

\usepackage{amsfonts}
\usepackage{graphicx}
\usepackage{slashbox}
\usepackage{dsfont}
\usepackage{mathtools}
\usepackage{hyperref}
\usepackage{url}
\usepackage{booktabs}
\usepackage[ruled,lined,commentsnumbered]{algorithm2e}
\usepackage{algorithmic}
\usepackage{amsmath,amssymb}
\usepackage{mathrsfs}

\IEEEoverridecommandlockouts                              

\title{\LARGE \bf
Learning Ball-balancing Robot Through Deep Reinforcement Learning
}

\author{Yifan~Zhou$^{1}$,
	  Jianghao~Lin$^{2}$,
      Shuai~Wang$^{3}$,
	  Chong~Zhang$^{3}$
\thanks{$^{1}$Y. Zhou is with Zhejiang University, Hangzhou, China
        (e-mail: yifanzhou7@gmail.com)}%
\thanks{$^{2}$J. Lin is with Shanghai Jiao Tong University, Shanghai, China
        (e-mail:chiangel@sjtu.edu.cn)}
\thanks{$^{3}$S. Wang and C. Zhang are with Tencent Robotics X, Shenzhen, China (e-mail: \{shawnshwang, chongzzhang\}@tencent.com)}

}

\begin{document}
\maketitle
\thispagestyle{empty}
\pagestyle{empty}

\begin{abstract}
The ball-balancing robot (ballbot) is a good platform to test the effectiveness of a balancing controller. Considering balancing control, conventional model-based feedback control methods have been widely used. However, contacts and collisions are difficult to model, and often lead to failure in balancing control, especially when the ballbot tilts a large angle. To explore the maximum initial tilting angle of the ballbot, the balancing control is interpreted as a recovery task using Reinforcement Learning (RL). RL is a powerful technique for systems that are difficult to model, because it allows an agent to learn policy by interacting with the environment. In this paper, by combining the conventional feedback controller with the RL method, a compound controller is proposed. We show the effectiveness of the compound controller by training an agent to successfully perform a recovery task involving contacts and collisions. Simulation results demonstrate that using the compound controller, the ballbot can keep balance under a larger set of initial tilting angles, compared to the conventional model-based controller. 

\textit{ }

\end{abstract}

\section{Introduction}
Ballbot is a dynamically-stable mobile robot balancing on a single spherical wheel. This unique way of moving makes ballbot move in omnidirectional and thus exceptionally agile and maneuverable compared to other ground vehicles. Also, continuously position adjusting allows robot to be tall enough for human-robot interaction and having large accelerations without tipping over. Balancing on a ball, apart from the impressive look and behavior, also makes the robot more applicable to human environments. The underlying dynamics of robotic system is highly nonlinear and unstable, which requires full system dynamics planning and control. While conventional feedback control method can solve task such as balancing or velocity control in omnidirectional direction, applications that involve contacts between the wheel and the ball, and between the ball and the ground are difficult to approach with conventional control method. We usually use pure viscous damping friction model, neglecting the static friction and nonlinear dynamic friction. Also, the large tilt angle will lead to the no-contact, collision and additional force, which cannot be modeled. Identifying and modeling contacts and friction under these circumstances is difficult, even if a reasonable contact behavior is provided by a carefully identified physical model. Hence, it is often difficult to achieve adaptable yet robust behavior as soon as the unmodelable elements are introduced.

RL algorithms have been widely used to learn policies for unstructured environments, because they allow agent to interact with their environments continuously and to iterate the policy spontaneously. Moreover, RL can handle control problems that are difficult to approach with conventional controllers because the control goal can be specified indirectly as a term in a reward function and not explicitly as the result of a control action  \cite{R1}. However, standard RL methods cannot be solely used in the ballbot, since it requires the robot learn through interaction, which can lead to instability. And collecting the amount of interaction that is needed to learn a complex skill can be time-consuming. 

In this paper, we attempt to combine the benefits of both conventional feedback control and RL. We study control problem that the ballbot have large initial tilt angle, which cannot be described by the model. The problems possess the modelable structure that can be partially handled with conventional feedback control. And the unmodelable part of the control task, which must consider contacts and friction, is solved with RL. 

The rest of the paper is organized as follows. In Section \ref{related}, related work on ballbot is reviewed and followed by the RL in control problems. Section \ref{pre} provide an overview of ballbot system and introduce the fundamental of RL. In Section \ref{method}, we present the control system. We give the dynamic modeling of the ballbot and design a feedback controller, followed by the introduction of the DDPG method. Section \ref{experiments} shows the simulation results implemented on ballbot, while Section \ref{discussion} concludes the paper. 

\section{Related Work}
\label{related}
\subsection{Ballbot}
\begin{figure}[tb]
      \centering
      \includegraphics[scale=0.33]{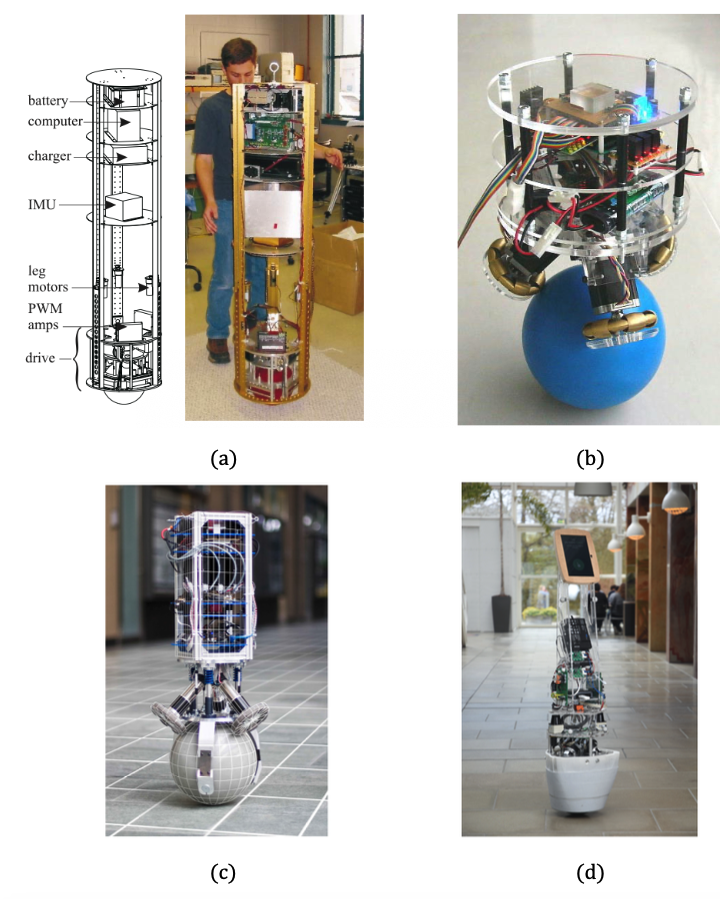}    
      \caption{Main ballbot contributions: (a) Ballbot from CMU; (b) ``BallIP" from TGU; (c) ``Rezero” from ETH (d) ``Kugle” from AAU.}
      \label{ballbot}
\end{figure}

Many ball-balancing robots have been built in the past years. The main ballbot research platform is shown in Fig. \ref{ballbot}. The CMU Ballbot \cite{CMU} used an inverse mouse-ball drive mechanism, consisting of four rollers to drive the ball. BallP \cite{BallP}, Rezero \cite{bookETH}, Kugle \cite{bookKugle} use three omni-wheels to drive the ball. A tall robot is built up for the convenience to interact with people. With a small leaning angle, the projection of the center of mass on the horizontal plane moves out of its supporting triangle. To increase the contact force or avoid the fact that the ball gets pushed out during large inclinations, the Kugle \cite{bookKugle} robot is equipped with a skirt and the Rezero \cite{bookETH} robot is equipped with ball arrester, which hold the ball in place. These equipment reduce the unmodelable elements' influence on the ballbot.

In  \cite{CMU}  \cite{BallP} and  \cite{bookETH}, linear feedback control approaches are used to maintain balance and achieve motion. The CMU ballbot uses an inner loop that maintains the body at desired inclination angles and an outer loop linear quadratic regulator (LQR) controller that achieves desired ball motions by commanding body angles to the inner controller. Later in  \cite{BallP} \cite{arm2}, CMU ballbot is introduced with arms and uses its planning procedure to plan in the space of both body lean angles and arm angles to achieve desired ball motions. The Kugle robot uses both linear feedback controllers \cite{bookKugle} and nonlinear sliding mode is designed to control the body angles to follow a given path.
\subsection{RL in Control Problems}
With RL getting increasingly popular, a large number of new complex robot control applications have been achieved (e.g., solving Rubik's Cube \cite{use1}, learning dexterous in-hand manipulation \cite{use2}, learning agile walk for legged robots \cite{use3}, and trajectory tracking of a rotorcraft aerial robot \cite{use5}). 

However, the dynamically stable robot is much less investigated due to the inherent instability which makes it impossible to collect enough data. And the random initial will easily cause robot tipping over or being destroyed. Hence, approaches that combine the convectional control method and the RL was proposed. A residual RL is used in robot arm control. The dynamical system consists of a fully actuated robot and underactuated objects. By decomposing them into a part that is solved efficiently by conventional feedback control methods, and the residual which is solved with RL, an agent was trained successfully to perform a real-world block assembly task involving contacts and unstable objects \cite{R4}. For the ballbot, a two-step algorithm is introduced which combines the proven performance of a model-based controller with a model-free method for compensating for model inaccuracy \cite{R3}. The authors build a linearized model of ballbot and design an iLQG controller, which can already stabilize the unstable system, enabling the learning algorithm to sample useful data. For the subsequent learning, the RL algorithm uses the designed iLQG controller as an optimized initial guess for the parameters for they share the same cost function. In this method, the RL is used to adapt the model-based controller to the real environment, rather than cope with the unmodelable problems.

\begin{figure}[tp]
      \centering
      \includegraphics[scale=0.36]{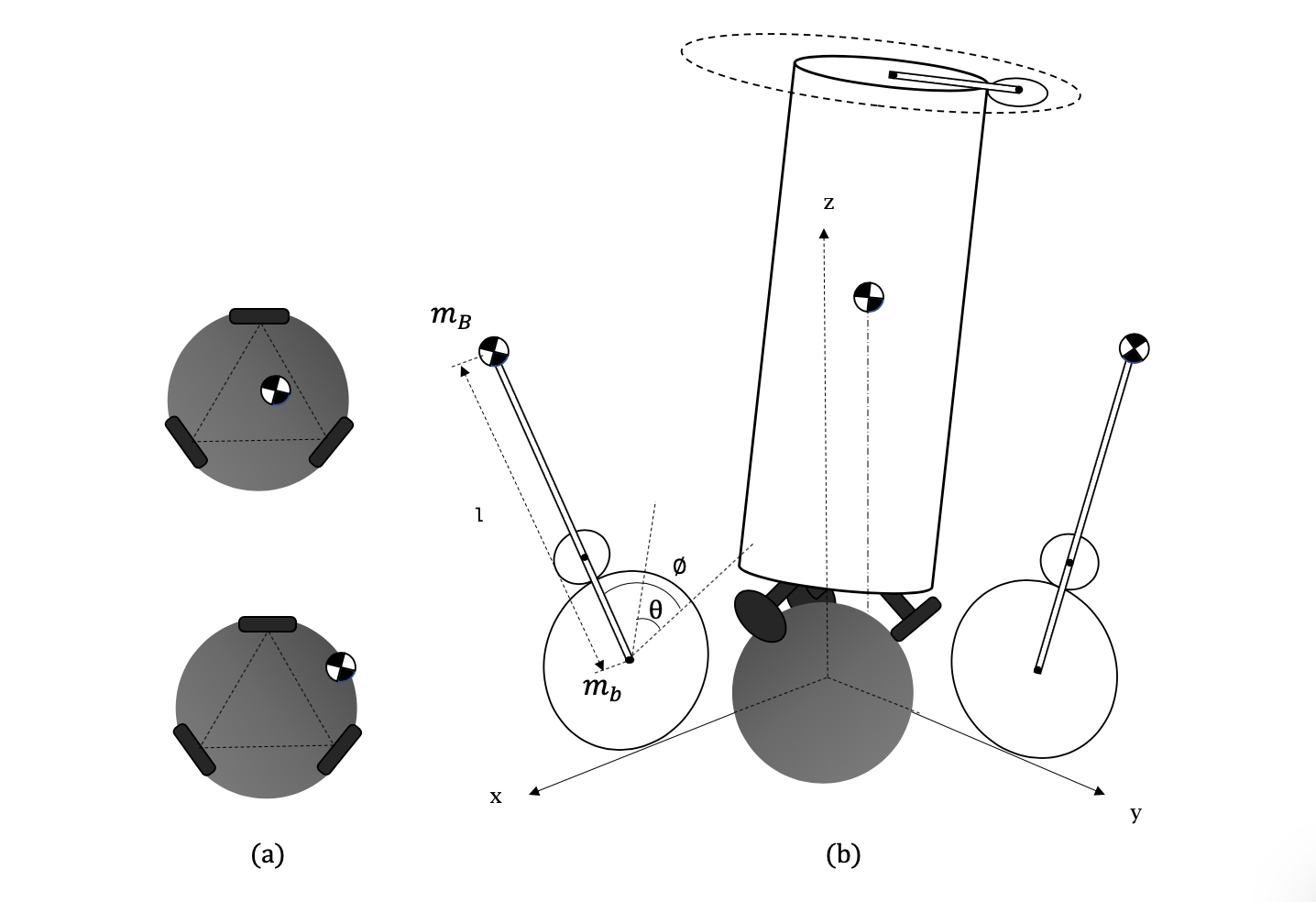}
      \caption{The model of the ballbot: (a) Top view of the ballbot; (b) Ballbot is approximated into two decoupled plant.}
      \label{model}
\end{figure}
\section{Preliminaries}
\label{pre}
In this section, we provide an overview of ballbot system. Also, the RL that we build on in our approach is introduced.

\subsection{System Overview}
The ballbot system consists of three rigid bodies: the ball, the actuating wheels and the body. The ball transmits and bears all arising forces, and the actuating wheels are three omniwheels mounted with $120^{\circ}$ spacing. Additionally, the following assumptions are made  \cite{CMU}:
\begin{itemize}
        \item \textit{Rigid Body Assumption}: In this model, all the objects are considered as rigid bodies, which means all the deformation should be neglected.
        
        \item \textit{Friction Assumption}: Friction between the wheel and the floor and between the wheel and the body is modeled as pure viscous damping. Forces due to static friction and nonlinear dynamic friction are neglected.
       
        \item \textit{No Slip Assumption}: There is no slip between the ball and floor and between the ball and the wheel.
\end{itemize}

The three contacting points on the ball form a supporting triangle, which is essential to the system. Consider two cases of the ballbot shown in Fig. \ref{model} (a). As the first case, the ballbot tilt a small angle and its center of gravity is directly above the supporting triangle, which guarantees that all wheels are contacting with the ball. However, for some reason, the inclination angles of a ballbot increase to a large extent, leading to the center of gravity out of the supporting triangle. With the gravity torque, the contact condition and the friction change a lot, thus the assumptions we made when modeling the robot are no longer satisfied. More specifically, there is a practical limit on the magnitude of the tilt angle. Exceeding this limit induces lots of unpredictable factors not present in the ballbot model. The unpredictable factors such as contact and friction in this model makes it impossible to identify the model accurately. Based on the discussion above, we consider the ballbot a system that consists of modelable part and unmodelable one.

\subsection{Reinforcement Learning}
In RL, we consider an agent interacting with an environment $E$ over discrete timesteps, which can be modeled as a Markov Decision Process. At every timestep $t$, the agent is in a state $s_t$, takes an action $a_t$, and the environment provides agent with a scalar reward $r_t$ and the state $s_{t+1}$. An agent’s behavior is defined by a policy $\pi$, which maps states to a probability distribution over actions $ \pi:\mathcal{S} \rightarrow \mathcal{P}(\mathcal{A})$. Policy learning can be subdivided into two categories: 1) Stochastic 2) Deterministic. Stochastic policy learns the conditional probability of taking an action $a$, given that it is in a state $s$, $\pi(a|s)=P[A_t=a|S_t=s]$. A deterministic policy on the other hand learns the action as a function of state, $a = \pi(s)$, which means that the agent will give a deterministic action in each state $s$.

We denote the return by $R_t=\sum_{i=t}^T \gamma^{(i-t)}r_i$, where $T$ is the horizon that the agent optimizes over and $\gamma \in [0,1]$ is a discount factor for future rewards. In RL, the agent aims to learn a policy $u_t=\pi(s_t)$ to maximize the expected return.

\begin{figure}[tb]
      \centering
      \includegraphics[width=\textwidth/2]{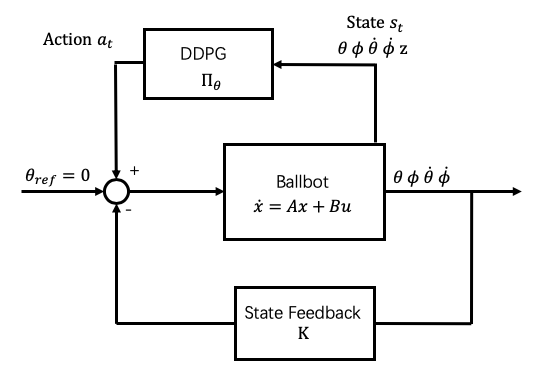}
      \caption{Structure of the system.}
      \label{system}
\end{figure}

\section{Balance Control}
\label{method}
Based on the analysis in Section \ref{pre}, we introduce a control system that consists of two parts. The first part is a conventional model-based linearized feedback controller that is able to keep the ballbot balance around the balancing point. It might have decreased performance on unmodelable conditions. And RL method is superposed and maximize the reward function we define. The control structure is illustrated in Fig. \ref{system}. In this way, the performance near the equilibrium will be guaranteed, and the iteration time may be decreased.

\subsection{Task Definition}
In this paper, we aim to design a controller that performs better than the conventional controller in the unmodelable section. To illustrate the ability of two controllers, we choose a balance recovery task, aiming to stabilize the unstable system from a tilt angle (with zero angular velocity) and keep balance for a period of time. 

\begin{figure}[tp]
      \centering
      \includegraphics[scale=0.35]{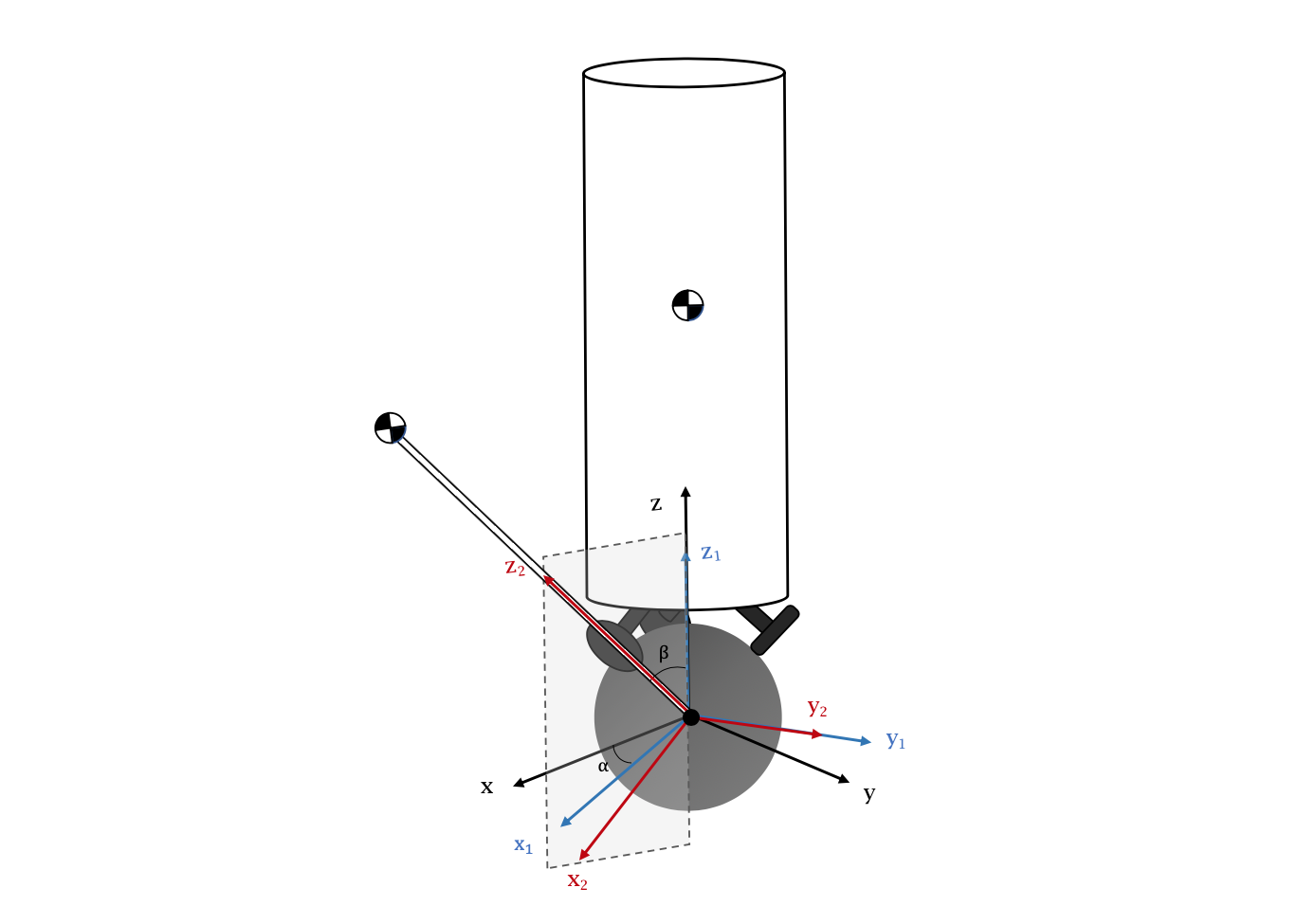}
      \caption{Parameters used to describe the initial state.}
      \label{orientation}
\end{figure}

Now we will give the parameters to describe the initial state of the robot. For the ballbot standing on the ball with all three omniwheels contacting with ball, the body can be regarded revolving around the center of the ball. As shown in Fig. \ref{orientation}, the orientation of the ball can be described: 1) rotates $\alpha$ around the z-axis of the ball 2) rotates $\beta$ around the y-axis of the ball. Generally speaking, the body has 2 DOFs: the inclination angle $\beta$ and the direction $\alpha$ to which the robot tilts.

\subsection{Conventional Feedback Controller}
\subsubsection{Simplified Mathematical Model}
A nonlinear 3D model of the system dynamics of ballbot has been analytically derived in \cite{CMU}, where the wheel dynamics are neglected. In this model, the three-dimensional system is approximated in three decoupled planes. As shown in Fig. \ref{model} (b), the median sagital plane (xz-plane) and the median coronal plane (yz-plane) related to the system moving forward or backward on the flat floor are identical and share same equations of motion. Wheels are simplified as one actuating wheel that fits in a plane. The third plane (xy-plane) describes the rotation around the z-axis in the body fixed reference frame, to which we pay less attention in balance control. Under these circumstances, the controller for the three-dimensional system can be designed by analysing and designing independent controllers for the two separate and identical systems.

We choose the median sagital plane to derive the model. In detail, the body angle $\theta$ represents the change of ball position calculated by $x$, which is read from sensors.
zhesh
where $r_b$ is the radius of the ball. Another configuration is $\phi$, aiming to describe the position of the robot above the ball. $\phi$ is accessed by IMU. The mass of the ballbot and the ball are given by $m_B$ and $m_b$ respectively. The distance from the center of mass of the ballbot to the center of mass of body is given by $l$.
The Lagrangian formulation is used to derive the nonlinear equations of motion for the simplified model. According to Euler-Lagrange equations, the energies of the system obeys:
\begin{equation}
\label{EulerLagrange}
\frac{d}{dt} \Big ( \frac{\partial \mathcal L}{\partial \dot{q}} \Big ) - \frac{\partial \mathcal L}{\partial q} = Q,
\end{equation}
where $\mathcal L$ is the difference between the kinetic energy and the potential energy, $q$ is generalized coordinate, $\dot q$ is the derivative of generalized coordinate $q$, and $Q$ is the external force, including the friction and the torque between the ball and the wheel.

The kinetic energy $K_b$ of the ball is computed with rotation energy and transnational kinetic energy:
\begin{equation}
\label{Kb}
K_b=\frac{I_b \dot\theta^2}{2}+\frac{m_b (r_b \dot\theta)^2}{2},
\end{equation}
where $I_b$, $m_b$, and $r_b$ are the moment of inertia, the mass and the radius of the ball respectively. The potential energy of the ball is $V_b=0$. The kinetic energy $K_B$ and potential energy $V_B$ of the body are
\begin{equation}
\begin{split}
K_B=&\frac{m_B}{2} (r_b^2 \dot\theta^2+2r_b l (\dot\theta^2+\dot\theta \dot\phi) \cos(\theta+\phi)+l^2 (\dot\theta+\dot\phi)^2) \\
&+\frac{I_B}{2} (\dot\theta +\dot\phi)^2,\\
V_B=&m_B g l \cos(\phi+\theta),
\end{split}
\end{equation}
where $I_B$ is the moment of inertia of the body with respect to the center of the ball, $m_B$ is the mass of the body, and $g$ is the gravity acceleration. The total kinetic energy is $K=K_b +K_B$ and the total potential energy is $V=V_b +V_B$.
Define the system configuration vector $q=\begin{bmatrix}\theta \;\; \phi \end{bmatrix}^T$. The Lagrangian $\mathcal L$ is a function of $q$ and $\dot q$ :
\begin{equation}
\mathcal{L}(q ,\dot{q})=K-V=K_B+K_b-V_B-V_b  .
\end{equation}
Let $\tau$ be the component of the torque applied between the ball and the body in the direction normal to the plane. To model the viscous friction terms, define the vector
\begin{equation}
D(\dot{q})=\begin{bmatrix} \mu_\theta \dot{\theta}\\ \mu_\phi \Dot{\phi} \end{bmatrix},
\end{equation}
where $\mu_\theta$ and $\mu_\phi$ are the viscous damping coefficients that model ball-ground and ball-body friction, respectively. Using this notation, the Euler-Lagrange equations of motion for the simplified ballbot model are
\begin{equation}
\frac{d}{dt} \Big ( \frac{\partial \mathcal L}{\partial \dot{q}} \Big ) - \frac{\partial \mathcal L}{\partial q} = \begin{bmatrix}\begin{matrix}
    0\\ \tau
    \end{matrix}\end{bmatrix}-D(\dot{q}).
\end{equation}
After computing the derivatives in the Euler-Lagrange equations and rearranging terms, the equations of motion can be expressed as
\begin{equation} 
 \label{eq1}
    M(q) \ddot{q}+C(q,\dot q)+G(q)+D(\dot q)=\begin{bmatrix}
    0 \\ \tau
    \end{bmatrix}.
\end{equation}
The mass matrix $M(q)$ is
\begin{equation}
M(q)=\begin{bmatrix}
\Gamma_1+2m_B r_b l \cos(\theta+\phi)&\Gamma_2+m_B r_b l \cos(\theta+\phi) \\
\Gamma_2+m_B r_b l \cos(\theta+\phi)&\Gamma_2
\end{bmatrix},
\end{equation}
where
\begin{equation}
\Gamma_1=I_b+I_B+m_b r_b^2+m_B r_b^2+m_B l^2,
\end{equation}
\begin{equation}
\Gamma_2=I_B+m_B l^2.
\end{equation}
The vector of Coriolis and centrifugal forces is
\begin{equation}
C(q,\dot{q})=\begin{bmatrix} -m_B r_b l \sin(\theta+\phi)(\dot{\theta}+\dot{\phi})^2 \\ 0 \end{bmatrix},
\end{equation}
and the vector of gravitational forces is
\begin{equation}
G(q)=\begin{bmatrix} -m_B g l \sin(\theta+\phi) \\  -m_B g l \sin(\theta+\phi) \end{bmatrix}.
\end{equation}
Premultiply both sides of (\ref{eq1}) by the inverse of matrix $M$:
\begin{equation}
\ddot{q}= M(q)^{-1} \Big(\begin{bmatrix}0 \\ \tau\end{bmatrix}-C(q,\dot{q})-G(q)-D(\dot{q})\Big).
\end{equation}
Rewrite the expression:
\begin{equation}
\ddot{q}= \begin{bmatrix}\ddot\theta \\ \ddot\phi \end{bmatrix}=\begin{bmatrix}\frac{b(\Gamma_2+l m_B r_b \cos(\phi +\theta))-a \Gamma_2}{c} \\
\frac{a (\Gamma_2 + l m_B r_b \cos(\phi +\theta)) -b (\Gamma_1 + 2 l m_B r_b \cos(\phi +\theta))}{c}
\end{bmatrix},
\end{equation}
where
\begin{equation}
\begin{split}
a=&m_B g l \sin(\phi +\theta)-\mu_\theta \dot{\theta}+l m_B r_b \sin(\phi +\theta){(\dot{\phi}+\dot{\theta})}^2 ,\\
b=&\tau - \mu_\phi \dot{\phi}+g l m_B \sin(\phi +\theta) ,\\
c=&\Gamma_2^2-\Gamma_1 \Gamma_2 + l^2 m_B^2 r_b^2 \cos^2(\phi +\theta). \\
\end{split}
\end{equation}

To put these equations into standard nonlinear state space form, define the state vector to be $x=\begin{bmatrix} q^T  \; \dot{q}^T \end{bmatrix}^T$ and define the input $u=\tau$. The relationship $\sin(\theta+\phi)=\theta+\phi$ and $\dot{\phi}=\dot{\theta}=0$ are satisfied when the ballbot system is at the equilibrium point. Then the linearized state space model can be described as:
\begin{equation}
\left\{
\begin{aligned}
\overset{.}x &=Ax+Bu \\
y &=Cx+Du ,\\
\end{aligned}
\right.
\end{equation}
where the inner state is
\begin{equation}
x = \begin{bmatrix}
\theta & \phi & \dot \theta & \dot \phi 
\end{bmatrix}^T,
\end{equation}
and the system input is
\begin{equation}
u = \tau
\end{equation}
with
\begin{equation}
A = \begin{bmatrix}
0 & 0 & 1 & 0 \\
0 & 0 & 0 & 1 \\
\frac{m_B^2 l^2 g r_b}{H} &\frac{m_B^2 l^2 g r_b}{H} & \frac{\mu_\theta\Gamma_2 }{H} & \frac{-\mu_\phi F}{H} \\
\frac{m_B g l(F-G)}{H} & \frac{m_B g l(F-G)}{H} & \frac{-\mu_{\theta}(\Gamma_2+m_b r_b l)}{H} & \frac{\mu_\phi G}{H}
\end{bmatrix},
\end{equation}
\begin{equation}
B = \begin{bmatrix}
0 & 0 & \frac{F}{H} & -\frac{G}{H}
\end{bmatrix}^T,
\end{equation}
where $F=\Gamma_2+ m_B r_b l$, $G= \Gamma+2 m_B r_b l$  and $H=\Gamma_2^2-\Gamma_1 \Gamma_2 + m_B^2 l^2 r_b^2$. $C$ and $D$ are the identity matrix and zero matrix with appropriate dimensions respectively.

\subsubsection{Design of Controller}
Now a linear state feedback controller can be used that stabilizes the system about $\phi=0$. We apply the state feedback control law
\begin{equation}
    u=-Kx
\end{equation}
to system, and we choose the structure of $K$ to be
\begin{equation}
    K=\begin{bmatrix}k_1 & k_2 &k_3 &k_4\end{bmatrix}.
\end{equation}
For the linearized system, a control law can be designed to ensure stability by traditional linear method.

\subsection{Balance Control Based on Deep Reinforcement Learning}
In this paper, we use DDPG algorithm \cite{DDPG}, a model-free, off-policy actor-critic algorithm using deep function approximators that provides us with a good framework to train the neural network for learning highly non-linear functions.

\subsubsection{Deep Deterministic Policy Gradient}
DDPG is an actor-critic policy learning method  \cite{actor_critic} with additional target networks to stabilize the learning process. The action-value function describing the expected return after taking an action $a_t$ in the state $s_t$:
\begin{equation}
\label{DDPG_1}
Q^\pi(s_t,a_t)=\mathbb{E}_{r_{i \geq t},s_{i>t} \sim E, a_i>t \sim \phi}[R_t|s_t,a_t],
\end{equation} and the recursive version of (\ref{DDPG_1}), known as the Bellman equation, is as follows:
\begin{equation}
Q^\pi(s_t,a_t)=\mathbb{E}_{r_t,s_{t+1} \sim E}[r(s_t,a_t)+\gamma \mathbb{E}_{a_{t+1} \sim \pi}[Q^\pi (s_{t+1},a_{t+1})]]
\end{equation}

DDPG is based on DQN \cite{DQN}, a commonly used off-policy algorithm. It uses the greedy policy $\pi(s) = \arg\max_{a}Q(s,a)$. We consider function approximators parameterized by $\theta^Q$, which we optimize by minimizing the loss:
\begin{equation}
    L(\theta^Q)=\mathbb{E}_{s_t \sim \rho^\beta,a_t \sim \beta,r_t \sim E}[(Q(s_t,a_t|\theta^Q)-y_t])^2,
\end{equation}
where
\begin{equation}
y_t=r(s_t,a_t)+\gamma Q(s_{t+1},\mu (s_{t+1})|\theta^Q).
\end{equation}

In DDPG algorithm the target policy is deterministic. We can describe it as a function $\mu: \mathcal{S} \rightarrow \mathcal{A}$ and avoid the inner expectation:
\begin{equation}
    Q^\mu(s_t,a_t)=\mathbb{E}_{r_t,s_{t+1} \sim E} \begin{bmatrix} r(s_t,a_t)+\gamma Q^\mu(s_{t+1},\mu(s_{t+1}))\end{bmatrix}
\end{equation}

The actor is updated by following the applying the chain rule to the expected return from the starting distribution $J$ with respect to the actor parameters:
\begin{equation}
\begin{split}
\nabla_{\theta^\mu}J\approx & \mathbb{E}_{s_t \sim \rho^\beta}[\nabla_{\theta^\mu}Q(s,a|\theta^Q)|_{s=s_t,a=\mu(s_t|\theta^\mu)}] \\
= & \mathbb{E}_{s_t \sim \rho^\beta}[\nabla_{a}Q(s,a|\theta^Q)|_{s=s_t,a=\mu(s_t)}\nabla_{\theta_\mu}\mu(s|\theta^\mu)|_{s=s_t}]
\end{split}
\end{equation}

To guarantee the samples' independence and identical distribution, a replay buffer $R$ is used, which also makes efficient use of hardware optimizations. Transitions are sampled from the environment according to the exploration policy and the tuple $(s_t, a_t, r_t, s_{t+1})$ is stored in the replay buffer. When this buffer becomes full, oldest samples are discarded. The actor and critic are updated by sampling a minibatch uniformly from the buffer. 

Additionally, DDPG treat the problem of exploration independently from the learning algorithm to deal with the in exploration in continuous action space. So the exploration policy $\mu'$ is constructed by adding noise sampled from a noise process $\mathcal{N}$ to the actor policy $\mu$
\begin{equation}
    \mu'(s_t)=\mu(s_t|\theta_t^\mu)+\mathcal{N}.
\end{equation}

When updating the target values, DDPG uses ``soft" target update, rather than directly copying the weights: $\theta' \leftarrow \tau \theta +(1-\tau)\theta'$ with $\tau \ll 1$, ensuring the consistently train without divergence. For more details, refer to Algorithm 1.

\begin{algorithm}
    \small\caption{DDPG algorithm}
    \label{alg:A}
    \begin{algorithmic}
    \STATE Randomly initialize critic network $Q(s,a|\theta^Q)$ and actor policy $\mu (s|\theta^{\mu})$ with weights $\theta^Q$ and $\theta^\mu$. \\
    \STATE Initialize target network $Q'$ and $\mu'$ with weights $\theta ^{Q'} \leftarrow \theta ^Q$, $\theta ^{\mu'} \leftarrow \theta ^\mu$
    \STATE Initialize relay buffer $R$
    \STATE \textbf{for} episode =$1 \rightarrow M$ \textbf{do} \\
        \quad Initialize a random process $\mathcal{N}$ for action exploration \\
        \quad Receive initial observation state $s_1$ \\
        \quad \textbf{for} $t=1 \rightarrow T$ \textbf{do} \\
        \quad\quad Select and execute action $a_t=\mu(s_t|\theta^\mu)+\mathcal{N}_t$ \\
        \quad\quad Observe reward $r_t$ and new state $s_{t+1}$ \\
        \quad\quad Store transition $(s_t,a_t,r_t,s_{t+1})$ in $R$ \\
        \quad\quad Sample a minibatch of transitions $(s_i,a_i,r_i,s_{i+1})$ from $R$ \\
        \quad\quad Set $y_i=r_i+\gamma Q' (s_{i+1},\mu(s_{i+1}|Q^{\mu'})|\theta^{Q'})$ \\
        \quad\quad Update critic network by minimizing the loss function: \\
        \begin{equation*}
            L=\frac{1}{N} \sum_{i}(y_i-Q(s_i,a_i|\theta^\mu))^2
        \end{equation*}
        \quad\quad Update actor policy using the sampled policy gradient:
        \begin{equation*}
        \nabla_{\theta_\mu}J\approx \frac{1}{N} \sum_{i}\nabla_a Q(s,a|\theta^Q)|_{s=s_i,s=\mu(s_i)}\nabla_{\theta^\mu}\mu(s|\theta^\mu)|_{s_i} 
        \end{equation*} 
        \quad\quad Update the target networks:
        \begin{equation*}
            \theta^{Q'}\leftarrow \tau \theta^Q + (1-\tau)\theta^{Q'}
        \end{equation*}
        \begin{equation*}
            \theta^{\mu'}\leftarrow \tau \theta^\mu + (1-\tau)\theta^{\mu'}
        \end{equation*}
        \quad\textbf{end for}\\
        \textbf{end for}
    \end{algorithmic}
    \end{algorithm}

\subsubsection{State Vector And Network Architecture}
Consistent with the conventional controller, the chosen state vector consists of the body angle $\phi$ and the angular velocity $\dot{\phi}$ in both the median sagital plane and the median coronal plane. As the large inclination angle will cause the problem that the ball might be pushed out, only this state cannot identify the full state of the ballbot and the ball. We introduce the the attitude of the robot $z$ to identify the full state of the ballbot.

And the action vector is the target velocity of the virtual wheel in each plane, which can be converted into the velocity for the real three motors. Considering the ability to improve the performance of the conventional controller, we decide the amplitude of the action approximately equals the max speed that the motor will reach.

The main object of our task is to provide a set of motor input that are needed to keep the ballbot in balance. The only element included in the reward function is $\beta$, the included angle between the z-axis in the body fixed reference frame and the vertical direction. The final reward function is as follows
\begin{equation}
    r_t=-||\beta||+c,
\end{equation}
where $c$ is a constant to make sure the reward function is always positive.

\section{Experiments}
\label{experiments}
We evaluate our method in simulation. In this section, we start by reporting how the experimental system is set up. Then the application of conventional feedback control method is detailed, followed by the improvement using DDPG algorithm. Several experiments are implemented, with results being analyzed.

\subsection{Experimental Setup}
\begin{figure}[tp]
      \centering
      \includegraphics[width=\textwidth/2]{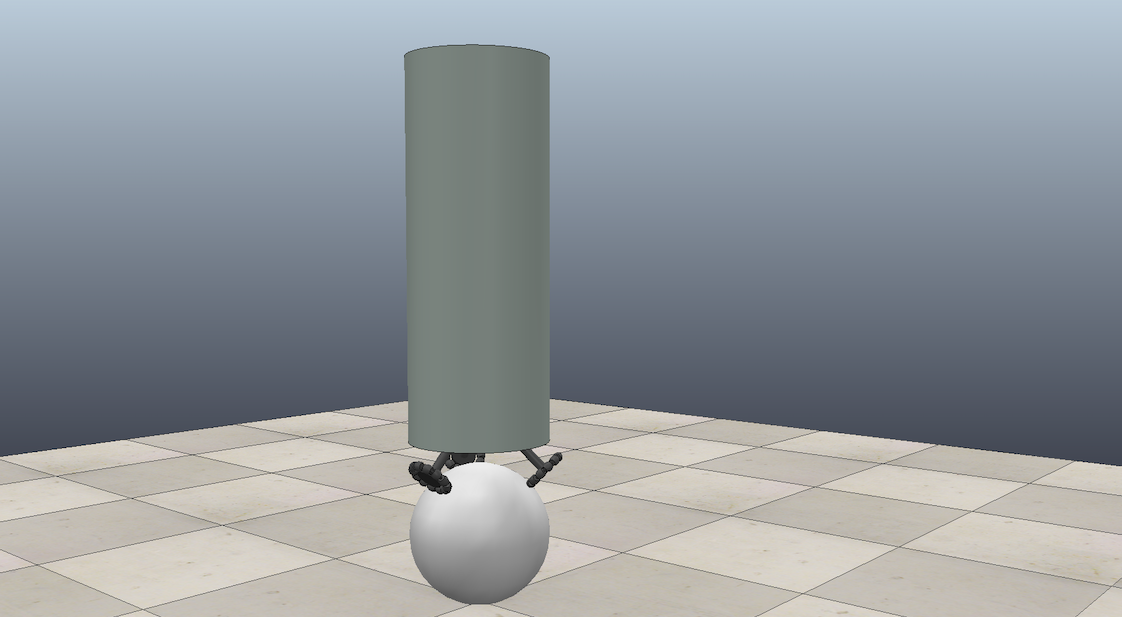}
      \caption{Ballbot in CoppeliaSim.}
      \label{vrep}
\end{figure}

We use CoppeliaSim $4.1.0$ shown in Fig. \ref{vrep}, a full-featured simulator for model-based optimization considering body contacts, to evaluate our method in simulation. This environment consists of a ballbot with three omni wheels, and without equipment such as skirt or ball arrester. The omni-wheel is driven by a motor controlled by a speed incremental controller. Also, we assume the full robot state information is available thanks to the virtual sensor on the body. To speed up the training process, we use the RL library, stable-baselines, which is an open-source library for reinforcement leaning that offers a simple interface to train, evaluate agents and do hyperparameter tuning for a variety of applications. 

The neural architectures for actor and critic network are both two fully connected Multi-Layer Perceptron with layer normalization. The other hyperparameters we used are listed in Table \ref{table}.

    \begin{table}[htpb]
    \centering
    \caption{HYPERPARAMETERS USED FOR DDPG}
    \label{table}
        \begin{tabular}{l l}
    \hline
            \textbf{Hyperparameter} & \textbf{Value} \\
    \hline
        	Hardware configuration & 1 CPU + 4 CPU cores \\
        	Optimizer & Adam \\
        	Actor learning rate & 0.0001 \\
        	Critic learning rate & 0.0001 \\
        	Weight decay regularization & 0.00001 \\
        	Minibatch size & 128 \\
        	Replay buffer size & 50000 \\
        	Constant $c$ in reward function & 0.7 \\
    \hline
        \end{tabular}
    \end{table}


\subsection{Experimental Results}

\begin{figure}[t]
      \centering
      \includegraphics[width=\textwidth/2]{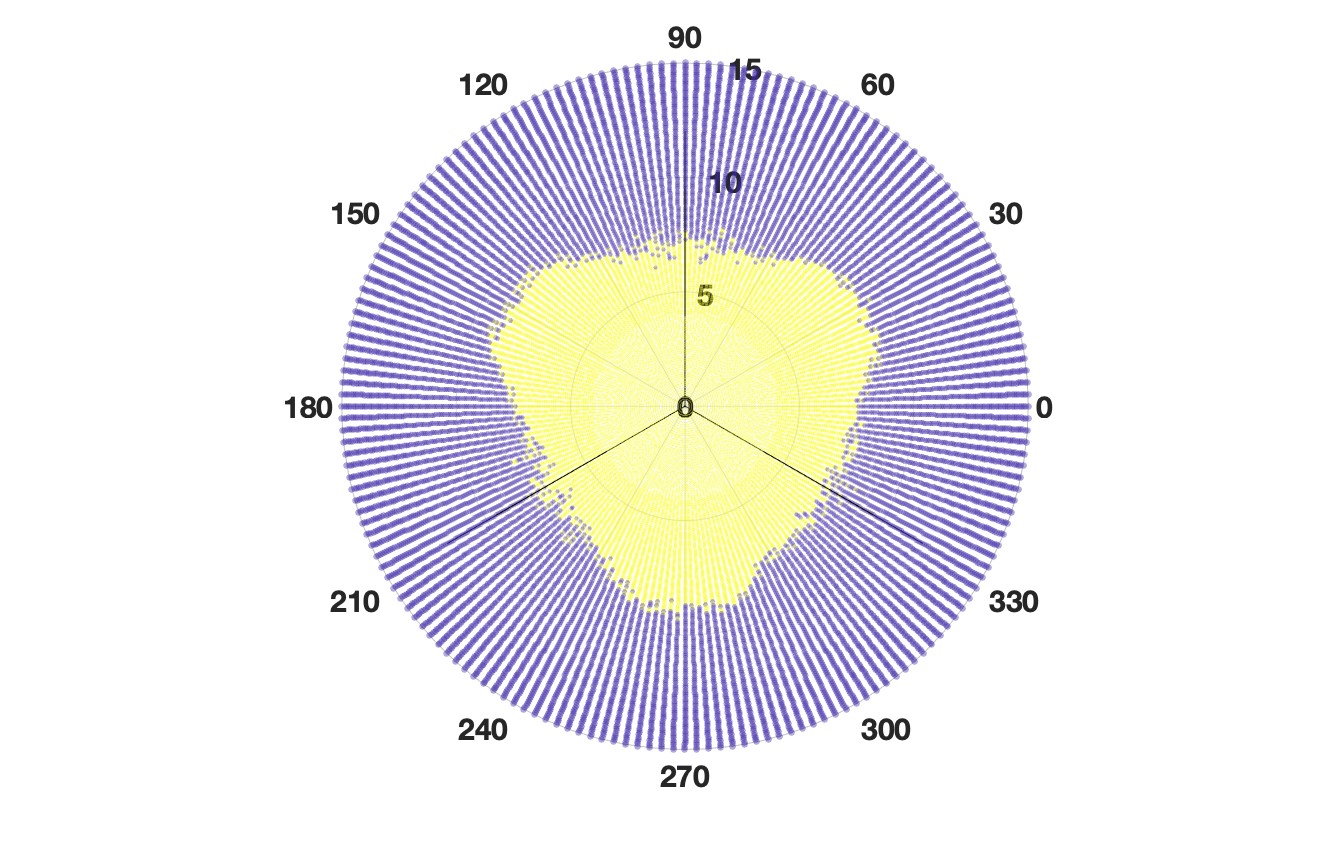}  
      \caption{The recovery area of the conventional controller.}
      \label{data1}
\end{figure}

\begin{figure}[t]
      \centering
      \includegraphics[width=\textwidth/2]{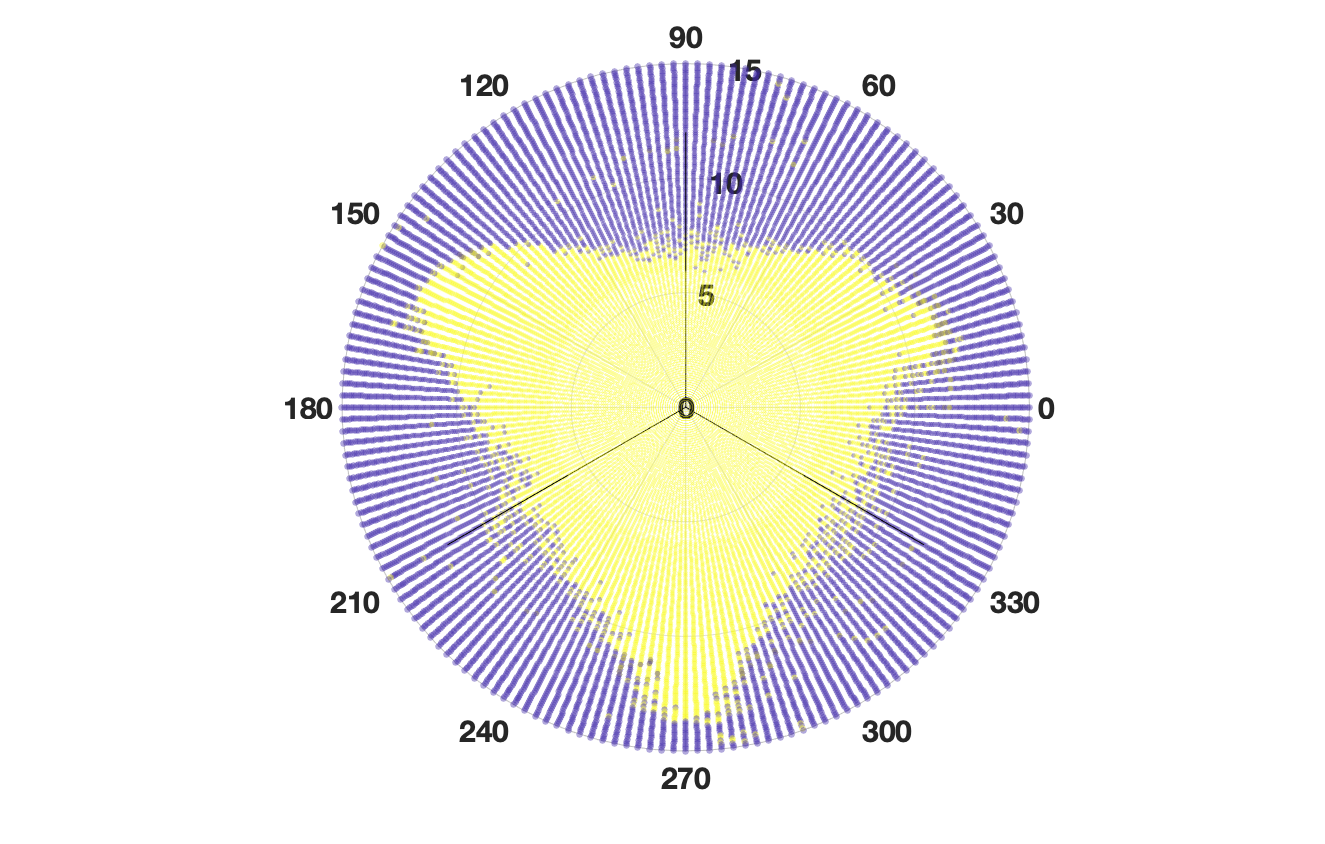}    
      \caption{The recovery area of the compound controller.}
      \label{data2}
\end{figure}

In the test run shown in Fig. \ref{data1} and Fig. \ref{data2}, a ballbot was commanded to recover from different initial states, with conventional controller and the compound controller respectively. The radial coordinate represents the angle $\beta$, and the angular coordinate represents the angle $\alpha$. All angles are expressed in degrees. Yellow points indicate that the ballbot can recover at those points, and purple points otherwise. The three lines mounted with $120^{\circ}$ spacing represent the wheel direction. Boundaries between yellow and purple points in both figures are roughly triangular. Define the yellow part as recovery area. 
Compared with the conventional controller, the compound controller has larger recoverable area. However, the ballbot has the least tilt angle $\beta$ with either conventional controller or the compound controller in the direction of the three wheels ($\alpha=90^{\circ}$, $210^{\circ}$ or $330^{\circ}$). The compound controller has very limited expending capacity at these directions. When the ballbot tilts a large angle in the direction $\alpha = 90^{\circ}$, only one omniwheel contacts with the ball. The force between the ball and the wheel is in the tangential direction.
The motor rotation contributes less to the balancing. When $\alpha=30^{\circ}$, $150^{\circ}$ or $270^{\circ}$,  the ballbot has two wheels contacting with the ball when $\alpha=30^{\circ}$, which allows the ball moving under two DOFs. Under these $\alpha$ angles, $\beta$ can be up to $12^{\circ} \sim 13^{\circ}$. Although the RL-based controller learns efficiently through contacting with the environment, it cannot break physical limitations of the system.

\begin{figure}[t]
      \centering
      \includegraphics[width=\textwidth/2]{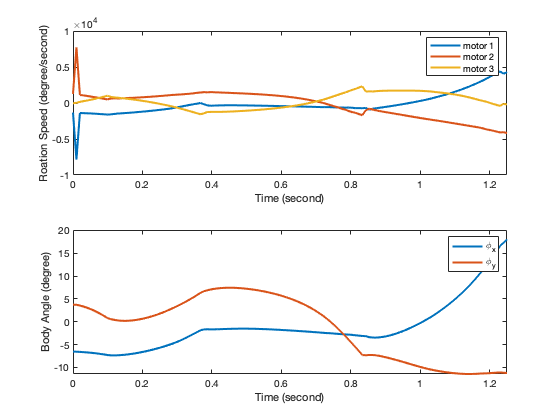}    
      \caption{The body angle $\phi_x$ ,$\phi_y$ and the rotation velocity of three motors with conventional controller in 1.25s.}
      \label{theta1}
\end{figure}

\begin{figure}[t]
      \centering
      \includegraphics[width=\textwidth/2]{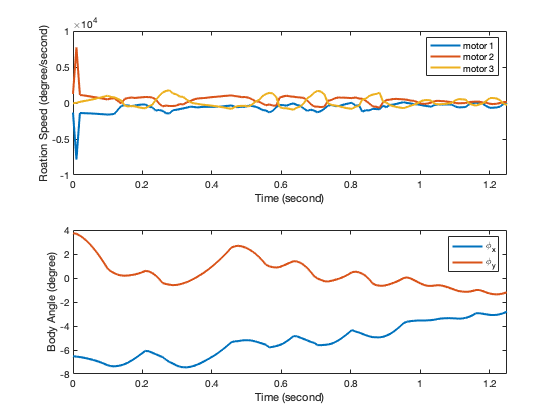}    
      \caption{The body angle $\phi_x$, $\phi_y$ and the rotation velocity of three motors with compound controller in 1.25s.}
      \label{theta2}
\end{figure}

Detailed states and inputs of the ballbot with respect to time are shown in  Fig. \ref{theta1} and Fig. \ref{theta2}. When $\alpha=210^{\circ}$, $\beta=7.5^{\circ}$, the ballbot with conventional controller falls over but the one with compound controller recovers. Body angles $\phi_x$, $\phi_y$ and the angular velocities of three motors are depicted. Oscillation exists during the adjusting time. Under this situation, the three wheels change their rotating directions rapidly and the ballbot “kicks” the ball to keep balance. Collision, bounce and slipping exist in this system, which are unpredictable and unmodelable.

\section{Conclusion}
\label{discussion}
In this paper, RL algorithm is superposed to the conventional controller, making up a compound controller. Both the conventional feedback controller and the compound controller can stabilize the ballbot system, while the compound controller has larger recovery area. 
When the compound controller tries to balance the ballbot from an initial state that the conventional controller fails, the ballbot ``kicks" the ball, which is an unmodelable but interesting phenomenon. This control policy can only be designed by RL. It is demonstrated that under certain circumstances, RL-based approach could give better or even unpredictable policies that can rarely be derived by conventional control framework.

\bibliographystyle{IEEEtran}
\bibliography{ieeeconf}

\end{document}